\documentclass[10pt,twocolumn,letterpaper]{article}

\usepackage[pagenumbers]{cvpr} %

\usepackage{graphicx}
\usepackage{amsmath}
\usepackage{amssymb}
\usepackage{booktabs}
\usepackage{makecell}
\usepackage[export]{adjustbox}
\usepackage[normalem]{ulem}
\usepackage{textpos}  %
\usepackage[misc]{ifsym}

\usepackage{tabulary,multirow,overpic,xcolor,subfloat}
\usepackage[pagebackref=false, breaklinks=true, letterpaper=true, colorlinks,
            citecolor=citecolor, linkcolor=linkcolor, bookmarks=false]{hyperref}
\definecolor{citecolor}{HTML}{0071BC}
\definecolor{linkcolor}{HTML}{ED1C24}

\newcommand{\app}{\raise.17ex\hbox{$\scriptstyle\sim$}}

\def\x{\times}
\newcolumntype{x}[1]{>{\centering\arraybackslash}p{#1pt}}
\newcolumntype{y}[1]{>{\raggedright\arraybackslash}p{#1pt}}

\newlength\savewidth\newcommand\shline{\noalign{\global\savewidth\arrayrulewidth
  \global\arrayrulewidth 1pt}\hline\noalign{\global\arrayrulewidth\savewidth}}
\newcommand{\tablestyle}[2]{\setlength{\tabcolsep}{#1}\renewcommand{\arraystretch}{#2}\centering\footnotesize}
\makeatletter\renewcommand\paragraph{\@startsection{paragraph}{4}{\z@}
  {.5em \@plus1ex \@minus.2ex}{-.5em}{\normalfont\normalsize\bfseries}}\makeatother

\newcommand\blfootnote[1]{\begingroup\renewcommand\thefootnote{}\footnote{#1}\addtocounter{footnote}{-1}\endgroup}

\DeclareMathAlphabet\mathbfcal{OMS}{cmsy}{b}{n}

\newcommand{\std}[1]{\fontsize{5pt}{0.1em}\selectfont #1}
\definecolor{Gray}{gray}{0.5}
\newcommand{\demph}[1]{\textcolor{Gray}{#1}}

\newcommand{\modelname}{Mask2Former\xspace}

\usepackage[capitalize]{cleveref}
\crefname{section}{Sec.}{Secs.}
\Crefname{section}{Section}{Sections}
\Crefname{table}{Table}{Tables}
\crefname{table}{Tab.}{Tabs.}

\newcommand{\tabref}[1]{Table~\ref{#1}}

\def\cvprPaperID{*****}
\def\confName{CVPR}
\def\confYear{2022}

\crefname{section}{\S}{\S\S}
\crefname{subsection}{\S}{\S\S}
\crefformat{table}{Table~#2#1#3}
\crefformat{figure}{Figure~#2#1#3}
\crefformat{equation}{Equation~(#2#1#3)}
\crefformat{algorithm}{Algorithm~#2#1#3}
\crefformat{appendix}{Appendix~#2#1#3}

\newcommand{\authorskip}{\hspace{2.5mm}}

\begin{document}

\title{\modelname for Video Instance Segmentation}

\author{
 B.~Cheng$^{1,2~{\textrm{\Letter}}}$ A.~Choudhuri$^1$ \authorskip I.~Misra$^2$ \authorskip
 A.~Kirillov$^2$ \authorskip R.~Girdhar$^{2^{\dagger}}$ \authorskip A.~G.~Schwing$^{1^{\dagger}}$ \\
 $^1$University of Illinois at Urbana-Champaign (UIUC) \authorskip $^2$Facebook AI Research (FAIR)\\
 {\small \url{https://github.com/facebookresearch/Mask2Former}}
}

\maketitle
\blfootnote{$^{\textrm{\Letter}}$ Work done while Bowen Cheng (\href{mailto:bcheng9@illinois.edu}{bcheng9@illinois.edu}) at FAIR.}
\blfootnote{$^\dagger$Equal advising.}

\begin{abstract}
We find \modelname~\cite{cheng2021mask2former} also achieves state-of-the-art performance on video instance segmentation \emph{without} modifying the architecture, the loss or even the training pipeline. In this report, we show universal image segmentation architectures trivially generalize to video segmentation by directly predicting 3D segmentation volumes. Specifically, \modelname sets a new state-of-the-art of 60.4 AP on YouTubeVIS-2019 and 52.6 AP on YouTubeVIS-2021. 
We believe \modelname is also capable of handling video semantic and panoptic segmentation, given its versatility in image segmentation. We hope this will make state-of-the-art video segmentation research more accessible and bring more attention to designing universal image and video segmentation architectures.

\end{abstract}

\section{Introduction}
Video instance segmentation~\cite{yang2019video} differs from  image segmentation in its goal to simultaneously segment \emph{and} track objects in videos.
Initial methods~\cite{yang2019video,voigtlaender2019mots} add  ``track embeddings'' to per-pixel image segmentation architectures to track objects across time. Recent transformer-based methods~\cite{vistr,ifc,wu2021seqformer} use ``object queries''~\cite{detr} to  connect objects across frames. However, all these methods are specifically designed to only process video data, disconnecting image and video segmentation research.

\begin{figure}
    \centering
    \includegraphics[width=\linewidth]{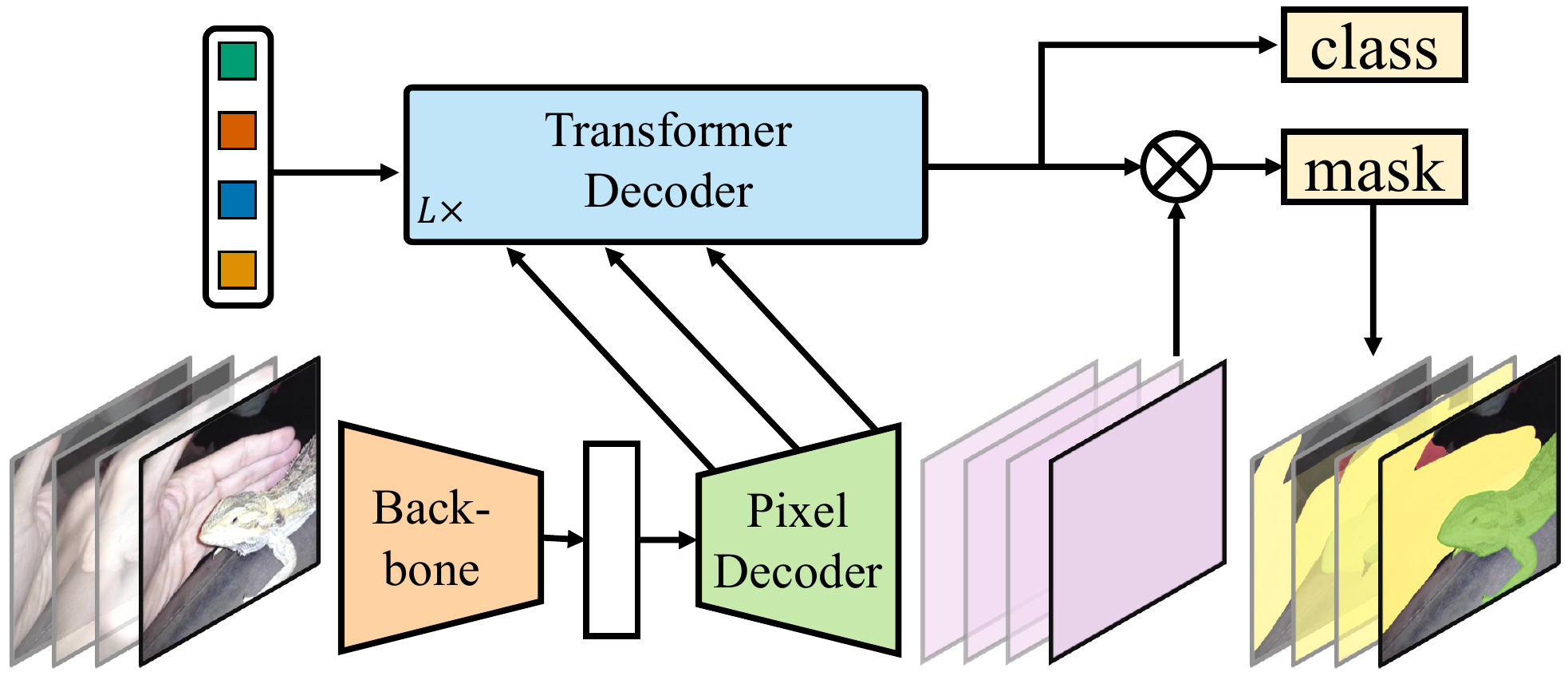}
    \caption{\modelname trivially generalizes to videos. For  single-frame input data, it operates as a standard image segmentation architecture. For $>1$ frames, due to sharing of the queries across  frames, it  segments and tracks  object instances across  frames.}
    \label{fig:teaser}
\end{figure}

Universal image segmentation methods~\cite{cheng2021maskformer,cheng2021mask2former} show that mask classification is able to address common image segmentation tasks (\ie panoptic, instance and semantic) using the same architecture, and achieve state-of-the-art results. 
A natural question emerges: does this universality extend to videos? In this report, we find  \modelname~\cite{cheng2021mask2former} achieves state-of-the-art results on video instance segmentation as well \emph{without} modifying the architecture, the loss or even the training pipeline. 
To achieve this, we let \modelname directly attend to the 3D spatio-temporal features and predict a 3D volume to track each object instance across time (Fig.~\ref{fig:teaser}). This simple change results in state-of-the-art performance of 60.4 AP on the challenging YouTubeVIS-2019 data and 52.6 AP on YouTubeVIS-2021.

\section{Background}
Video instance segmentation (VIS)~\cite{yang2019video} requires tracking instances across time. In general, there are two ways to extend image segmentation models for the VIS task:

\noindent\textbf{Per-frame} methods (\aka \textbf{online} methods) treat a video clip as a sequence of frames. They run image segmentation models on each frame independently and associate predicted instance masks across frames with a post-processing step. MaskTrack R-CNN~\cite{yang2019video}, TrackR-CNN~\cite{voigtlaender2019mots} and VPSNet~\cite{kim2020video} add per-instance track embedding prediction to Mask R-CNN~\cite{he2017mask}. VIP-DeepLab~\cite{qiao2021vip} extends Panoptic-DeepLab~\cite{cheng2020panoptic} by predicting instance center offsets across frames. 
However, it is not always straightforward to modify an image segmentation model, as architectural changes and additional losses are needed.

\noindent\textbf{Per-clip} methods (\aka \textbf{offline} methods) treat a video clip as a 3D spatio-temporal volume and directly predict the 3D mask for each instance. STEm-Seg~\cite{athar2020stem} predicts and clusters spatio-temporal instance embeddings to generate 3D masks. More recently, inspired by the success of DETR~\cite{detr}, VisTR~\cite{vistr}, IFC~\cite{ifc} and SeqFormer~\cite{wu2021seqformer} design Transformer-based architectures to process the 3D volume via cross-attention. However, all these models are designed specifically for video instance segmentation, disconnecting image and video segmentation research. In this report, we show Mask2Former~\cite{cheng2021mask2former}, a state-of-the-art universal image segmentation model, can also achieve state-of-the-art video instance segmentation \emph{without} modifying the architecture, the loss or even the training pipeline.

\section{\modelname for Videos}
We treat a video sequence as a 3D spatio-temporal volume of dimension $T \x H \x W$, where $T$ is the number of frames and $H, W$ are height and width respectively. We make three changes to adapt \modelname  to video segmentation: 1) we apply the masked attention to the spatio-temporal volume; 2) we add an extra positional encoding for the temporal dimension; and 3) we directly predict a 3D volume of an instance across time.

\subsection{Joint spatio-temporal masked attention}
We apply masked attention to the 3D spatio-temporal features, \ie, we use
\begin{align}
  \mathbf{X}_{l} = \text{softmax}(\mathbfcal{M}_{l-1} + \mathbf{Q}_{l}\mathbf{K}^{\text{T}}_{l})\mathbf{V}_{l} + \mathbf{X}_{l-1}.
\end{align}
Here, $l$ is the layer index, $\mathbf{X}_{l} \in \mathbb{R}^{N \x C}$ refers to $N$ $C$-dimensional query features at the $l^{\text{th}}$ layer and $\mathbf{Q}_{l} = f_Q(\mathbf{X}_{l-1}) \in \mathbb{R}^{N \x C}$. $\mathbf{X}_{0}$ denotes input query features of the Transformer decoder. $\mathbf{K}_{l}, \mathbf{V}_{l} \in \mathbb{R}^{TH_{l}W_{l} \x C}$ are the spatio-temporal features under transformation $f_K(\cdot)$ and $f_V(\cdot)$ respectively, $T$ is the number of frames and $H_l$ and $W_l$ are the spatial resolution. $f_Q$, $f_K$ and $f_V$ are linear functions.

Moreover, the 3D attention mask $\mathbfcal{M}_{l-1}$ at feature location $(t, x, y)$ is
\begin{align}
\mathbfcal{M}_{l-1}(t, x, y) = \left\{\begin{array}{ll}
  0  & \text{if~} \mathbf{M}_{l-1}(t, x,y)=1 \\
    -\infty & \text{otherwise}
\end{array}\right..
\end{align}
Here, $\mathbf{M}_{l-1} \in \{0,1\}^{N \x TH_{l}W_{l}}$ is the binarized output (thresholded at $0.5$) of the resized 3D mask prediction of the previous ($l-1$)-th Transformer decoder layer.

\subsection{Temporal positional encoding}
To obtain compatibility with image segmentation models we decouple the temporal positional encoding from the spatial  encoding, \ie, we use the positional encoding
\begin{align}
  e_{\text{pos}} = e_{\text{pos-t}} \oplus e_{\text{pos-xy}}.
\end{align}
Here, $e_{\text{pos}} \in \mathbb{R}^{T \x H_{l} \x W_{l} \x C}$ is the final positional encoding, $e_{\text{pos-t}} \in \mathbb{R}^{T \x 1 \x 1 \x C}$ and $e_{\text{pos-xy}} \in \mathbb{R}^{1 \x H_{l} \x W_{l} \x C}$ are the corresponding temporal and spatial positional encodings. Both $e_{\text{pos-t}}$ and $e_{\text{pos-xy}}$ are non-parametric sinusoidal positional encodings that can handle arbitrary length. $\oplus$ denotes addition with \texttt{numpy}-style broadcasting.

\subsection{Joint spatio-temporal mask prediction}

Similarly to the masked attention, we obtain the 3D mask of the $n^{\text{th}}$ query via a simple dot product, \ie, 
\begin{align}
\mathbf{M}(n, t, h, w) = \text{sigmoid}(\mathbf{E}_\text{mask}(:, n)^{\text{T}} \cdot \mathbf{E}_\text{pixel}(:, t, h, w)).
\end{align}
Note, computation of the classification  does not change.

\section{Experiments}
We evaluation \modelname on YouTubeVIS-2019 and YouTubeVIS-2021~\cite{yang2019video}. We do not modify the architecture, the loss or even the training pipeline.

\begin{table}[t]
  \centering
  \tablestyle{3.5pt}{1.2}
  \scriptsize
  \begin{tabular}{c|l|ll|x{28}x{24}x{24}}
  & method & backbone & data & AP & AP50 & AP75 \\
  \shline
  \multirow{9}{*}{\rotatebox{90}{CNN}} &
  \multirow{2}{*}{VisTR~\cite{vistr}} & R50 & V & 36.2 \phantom{\std{$\pm$0.5}} & 59.8 & 36.9 \\
  & & R101 & V & 40.1 \phantom{\std{$\pm$0.5}} & 45.0 & 38.3 \\
  \cline{2-7}
  & \multirow{2}{*}{IFC~\cite{ifc}} & R50 & V & 41.2 \phantom{\std{$\pm$0.5}} & 65.1 & 44.6 \\
  & & R101 & V & 42.6 \phantom{\std{$\pm$0.5}} & 66.6 & 46.3 \\
  \cline{2-7}
  & \multirow{3}{*}{SeqFormer~\cite{wu2021seqformer}} & R50 & V & 45.1 \phantom{\std{$\pm$0.5}} & 66.9 & 50.5 \\
  & & R50 & V + C80k & 47.4 \phantom{\std{$\pm$0.5}} & 69.8 & 51.8 \\
  & & R101 & V + C80k & 49.0 \phantom{\std{$\pm$0.5}} & 71.1 & \textbf{55.7} \\
  \cline{2-7}
  & \multirow{2}{*}{\textbf{\modelname}} & R50 & V & 46.4 \std{$\pm$0.8} & 68.0 & 50.0 \\
  & & R101 & V & \textbf{49.2} \std{$\pm$0.7} & \textbf{72.8} & 54.2 \\
  \hline\hline
  \multirow{6}{*}{\rotatebox{90}{Transformer}} &
  SeqFormer~\cite{wu2021seqformer} & Swin-L & V + C80k & 59.3 \phantom{\std{$\pm$0.5}} & 82.1 & 66.4 \\
  \cline{2-7}
  & \multirow{4}{*}{\textbf{\modelname}} & Swin-T & V & 51.5 \std{$\pm$0.7} & 75.0 & 56.5 \\
  & & Swin-S & V & 54.3 \std{$\pm$0.7} & 79.0 & 58.8 \\
  & & Swin-B & V & 59.5 \std{$\pm$0.7} & 84.3 & 67.2 \\
  & & Swin-L & V & \textbf{60.4} \std{$\pm$0.5} & \textbf{84.4} & \textbf{67.0} \\
  \cline{2-7}
  & \demph{best of 5 runs} & \demph{Swin-L} & V & \demph{60.7} \phantom{\std{$\pm$0.5}} & \demph{84.4} & \demph{66.7} \\
  \end{tabular}

  \caption{
  \textbf{YouTubeVIS-2019 \texttt{val}.} \modelname outperforms all state-of-the-art models \emph{without} using image data for augmentation. \textit{V:} YouTubeVIS-2019 \texttt{train} set. \textit{C80k:} 80k COCO \texttt{train2017} images that contain YouTubeVIS categories. We report the median of 5 runs together with the standard deviation.
  }

\label{tab:results:ytvis2019}
\end{table}

\begin{table}[t]
  \centering
  \tablestyle{3.5pt}{1.2}
  \scriptsize
  \begin{tabular}{c|l|ll|x{28}x{24}x{24}}
  & method & backbone & data & AP & AP50 & AP75 \\
  \shline
  \multirow{4}{*}{\rotatebox{90}{CNN}} &
  IFC~\cite{ifc} & R50 & V & 36.6 \phantom{\std{$\pm$0.5}} & 57.9 & 39.3 \\
  & SeqFormer~\cite{wu2021seqformer} & R50 & V + C80k & 40.5 \phantom{\std{$\pm$0.5}} & 62.4 & 43.7 \\
  \cline{2-7}
  & \multirow{2}{*}{\textbf{\modelname}} & R50 & V & 40.6 \std{$\pm$0.7} & 60.9 & 41.8 \\
  & & R101 & V & \textbf{42.4} \std{$\pm$0.6} & \textbf{65.9} & \textbf{45.8} \\
  \hline\hline
  \multirow{6}{*}{\rotatebox{90}{Transformer}} &
  SeqFormer~\cite{wu2021seqformer} & Swin-L & V + C80k & 51.8 \phantom{\std{$\pm$0.5}} & 74.6 & 58.2 \\
  \cline{2-7}
  & \multirow{4}{*}{\textbf{\modelname}} 
  & Swin-T & V & 45.9 \std{$\pm$0.6} & 68.7 & 50.7 \\
  & & Swin-S & V & 48.6 \std{$\pm$0.4} & 77.2 & 52.0 \\
  & & Swin-B & V & 52.0 \std{$\pm$0.6} & 76.5 & 54.2 \\
  & & Swin-L & V & \textbf{52.6} \std{$\pm$0.7} & \textbf{76.4} & \textbf{57.2} \\
  \cline{2-7}
  & \demph{best of 5 runs} & \demph{Swin-L} & V & \demph{53.0} \phantom{\std{$\pm$0.5}} & \demph{75.9} & \demph{58.4} \\
  \end{tabular}

  \caption{
  \textbf{YouTubeVIS-2021 \texttt{val}.} \modelname outperforms all state-of-the-art models \emph{without} using image data for augmentation. \textit{V:} YouTubeVIS-2021 \texttt{train} set. \textit{C80k:} 80k COCO \texttt{train2017} images that contain YouTubeVIS categories. For Swin-L models, we evaluate with a lower resolution (shorter side to 360 pixels) due to memory constraints. We report the median of 5 runs together with the standard deviation.
  }

\label{tab:results:ytvis2021}
\end{table}

\subsection{Training}
We use Detectron2~\cite{wu2019detectron2} and follow the IFC~\cite{ifc} settings for video instance segmentation. More specifically, we use the  AdamW~\cite{loshchilov2018decoupled} optimizer and the step learning rate schedule. We use an initial learning rate of $0.0001$ and a weight decay of $0.05$ for all backbones. A learning rate multiplier of $0.1$ is applied to the backbone and we decay the learning rate at $2/3$ fractions of the total number of training steps by a factor of $10$. We train our models for $6$k iterations with a batch size of $16$ for YouTubeVIS-2019 and $8$k iterations for YouTubeVIS-2021. During training, each video clip is composed of $T = 2$ frames, which makes the training much more efficient (2 hours per training with 8 V100 GPUs),  and the shorter spatial side is resized to either 360 or 480. All models are initialized with COCO~\cite{lin2014coco} instance segmentation models from~\cite{cheng2021mask2former}. Unlike~\cite{wu2021seqformer}, we only use YouTubeVIS training data and do not use COCO images for data augmentation. 

\subsection{Inference}

Our model is able to handle video sequences of various length. During inference, we provide the whole video sequence as input to the model and obtain 3D mask predictions without any post-processing. We keep the top 10 predictions for each video sequence. If not stated otherwise, we resize the shorter side to 360 pixels during inference for ResNet~\cite{he2016deep} backbones and to 480 pixels for Swin Transformer~\cite{liu2021swin} backbones.

\subsection{Results}

We compare \modelname with state-of-the-art models on the YouTubeVIS-2019 dataset in \tabref{tab:results:ytvis2019} and the YouTubeVIS-2021 dataset in \tabref{tab:results:ytvis2021}. Using the exact same training parameters, \modelname outperforms IFC~\cite{ifc} by more than 6~AP. \modelname also outperforms concurrent SeqFormer~\cite{wu2021seqformer} without using extra COCO images for data augmentation.

\paragraph{Acknowledgments:} B.~Cheng, A.~Choudhuri and A.~Schwing were supported in part by NSF grants \#1718221, 2008387, 2045586, 2106825, MRI \#1725729, NIFA award 2020-67021-32799 and Cisco Systems Inc.\ (Gift Award CG 1377144 - thanks for access to Arcetri).

{\small
\bibliographystyle{ieee_fullname}
\bibliography{refs}
}

\end{document}